\documentclass[letterpaper, 10 pt, conference]{ieeeconf}  

\IEEEoverridecommandlockouts
\usepackage{xcolor}
\usepackage{graphicx}
\usepackage{caption}
\usepackage{subcaption}
\usepackage{amsmath}
\usepackage{algorithm}
\usepackage{algorithmicx}
\usepackage[noend]{algpseudocode}
\usepackage{verbatim} 
\usepackage[font={footnotesize}]{caption}
\usepackage{amssymb}

\algrenewcommand\algorithmicrequire{\textbf{Input:}}
\algrenewcommand\algorithmicensure{\textbf{Output:}}
\errorcontextlines\maxdimen

\makeatletter
\newcommand*{\algrule}[1][\algorithmicindent]{\makebox[#1][l]{\hspace*{.5em}\vrule height .75\baselineskip depth .25\baselineskip}}%

\newcount\ALG@printindent@tempcnta
\def\ALG@printindent{%
    \ifnum \theALG@nested>0
        \ifx\ALG@text\ALG@x@notext
            \addvspace{-3pt}
        \else
            \unskip
            \ALG@printindent@tempcnta=1
            \loop
                \algrule[\csname ALG@ind@\the\ALG@printindent@tempcnta\endcsname]%
                \advance \ALG@printindent@tempcnta 1
            \ifnum \ALG@printindent@tempcnta<\numexpr\theALG@nested+1\relax
            \repeat
        \fi
    \fi
    }%
\usepackage{etoolbox}
\patchcmd{\ALG@doentity}{\noindent\hskip\ALG@tlm}{\ALG@printindent}{}{\errmessage{failed to patch}}
\makeatother

\makeatletter
\newcommand\fs@nobottomruled{\def\@fs@cfont{\bfseries}\let\@fs@capt\floatc@ruled
  \def\@fs@pre{\hrule height.8pt depth0pt \kern2pt}%
  \def\@fs@post{}
  \def\@fs@mid{\kern2pt\hrule\kern2pt}%
  \let\@fs@iftopcapt\iftrue}
\makeatother

\floatstyle{nobottomruled}
\restylefloat{algorithm}

\title{\LARGE \bf
Real-Time Planning Under Uncertainty for AUVs Using Virtual Maps
\vspace{-4mm}}
\author{Ivana Collado-Gonzalez,
        John McConnell, Jinkun Wang, Paul Szenher, and Brendan Englot \vspace{-3mm}
\thanks{\footnotesize{I. Collado-Gonzalez, P. Szenher, and B. Englot are with Stevens Institute of Technology, Hoboken, NJ, USA, \{\texttt{icollado}, \texttt{pszenher}, \texttt{benglot}\}\texttt{@stevens.edu}. J. McConnell was previously with Stevens and is now with the United States Naval Academy, Annapolis, MD, USA, \texttt{jmcconne@usna.edu}. J. Wang was previously with Stevens and is now with Anyware Robotics, Fremont, CA USA. \textbf{This research was supported by NSF Grant IIS-1652064 and USDA-NIFA Grant 2021-67022-35977.}}}
}

\begin{document}
\maketitle
\thispagestyle{empty}
\pagestyle{empty}

\begin{abstract} 
Reliable localization is an essential capability for 
marine robots navigating in GPS-denied 
environments. SLAM, commonly used to mitigate dead reckoning errors, still fails in feature-sparse environments or with limited-range sensors. Pose estimation can be improved by incorporating the uncertainty prediction of future poses into the planning process and choosing actions that reduce uncertainty. However, performing belief propagation is computationally costly, especially when operating in large-scale environments. This work proposes a computationally efficient planning under uncertainty framework suitable for large-scale, feature-sparse environments. Our strategy leverages SLAM graph and occupancy map data obtained from a prior exploration phase to create a \textit{virtual map}, describing the uncertainty of each map cell using a multivariate Gaussian. 
The virtual map is then used as a cost map in the planning phase, and performing belief propagation at each step is avoided. A receding horizon 
planning strategy is implemented, managing a goal-reaching and uncertainty-reduction tradeoff. 
Simulation experiments in a realistic underwater environment validate this approach. Experimental comparisons against a full belief propagation approach and a standard shortest-distance approach are conducted. 

\end{abstract}

\vspace{-2mm}

\section{Introduction}


\vspace{-1mm}

Autonomous Underwater Vehicles (AUVs) are being increasingly used to perform 
tasks such as inspection, mapping, 
and exploration. 
Since GPS is unavailable underwater, AUVs must rely on onboard sensors to estimate their position, making them susceptible to drift. Simultaneous Localization and Mapping (SLAM) has been successfully applied to correct dead reckoning drift. However, SLAM requires environmental features to lie within the AUV's field of view. Therefore, robots operating in feature-sparse 
environments or with short-range sensors are especially prone to accumulating pose estimation errors. Thus, executing trajectories that do not account for robot sensing capabilities leads to high localization uncertainty. Planning under uncertainty, however, commonly relies on belief propagation, which is computationally expensive and can become infeasible to compute when operating in large-scale environments. 

\begin{figure}[th]
    \centering
    \begin{subfigure}[b]{0.535\textwidth}
        \centering
        \rotatebox[origin=c]{0}{\includegraphics[scale = 0.29]
        {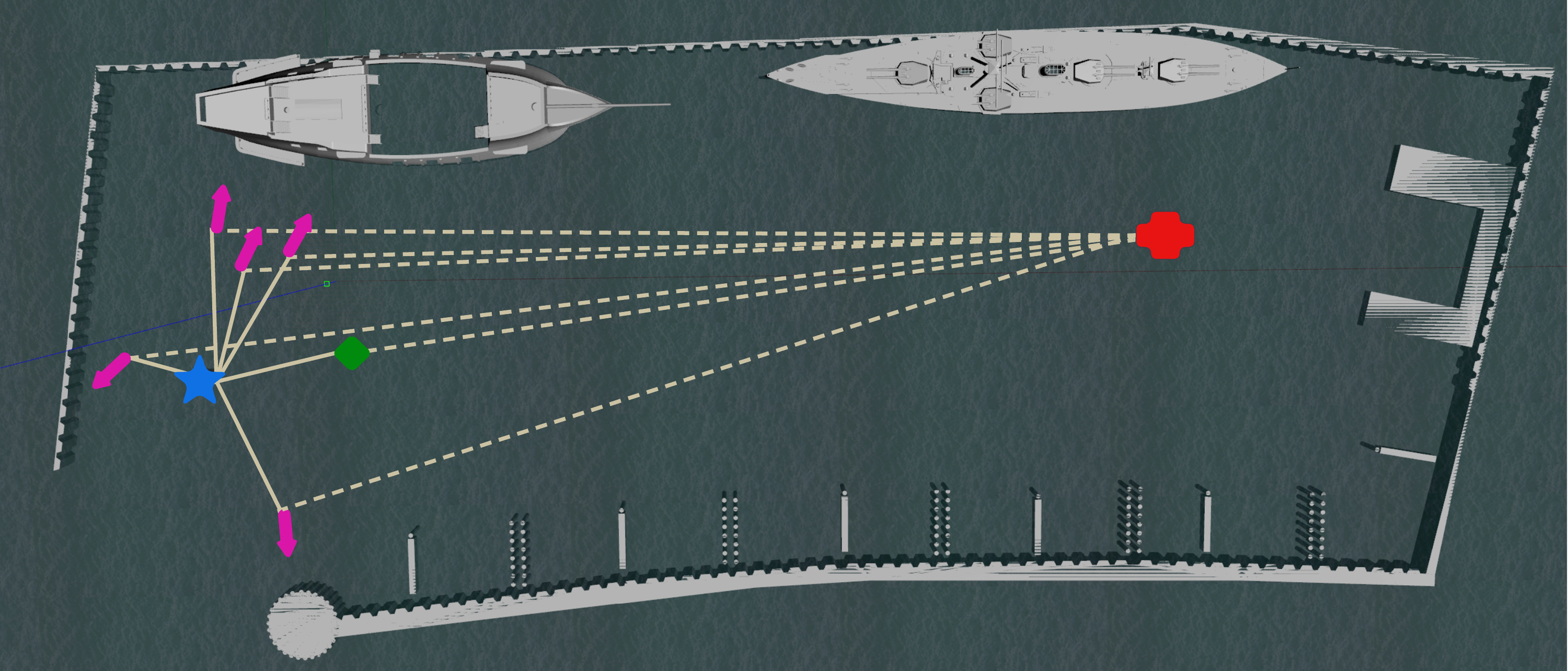}}
        \label{fig:gazebo}
    \end{subfigure}
    \begin{subfigure}[b]{0.5\textwidth}
        \centering
        \includegraphics[scale = 0.12]{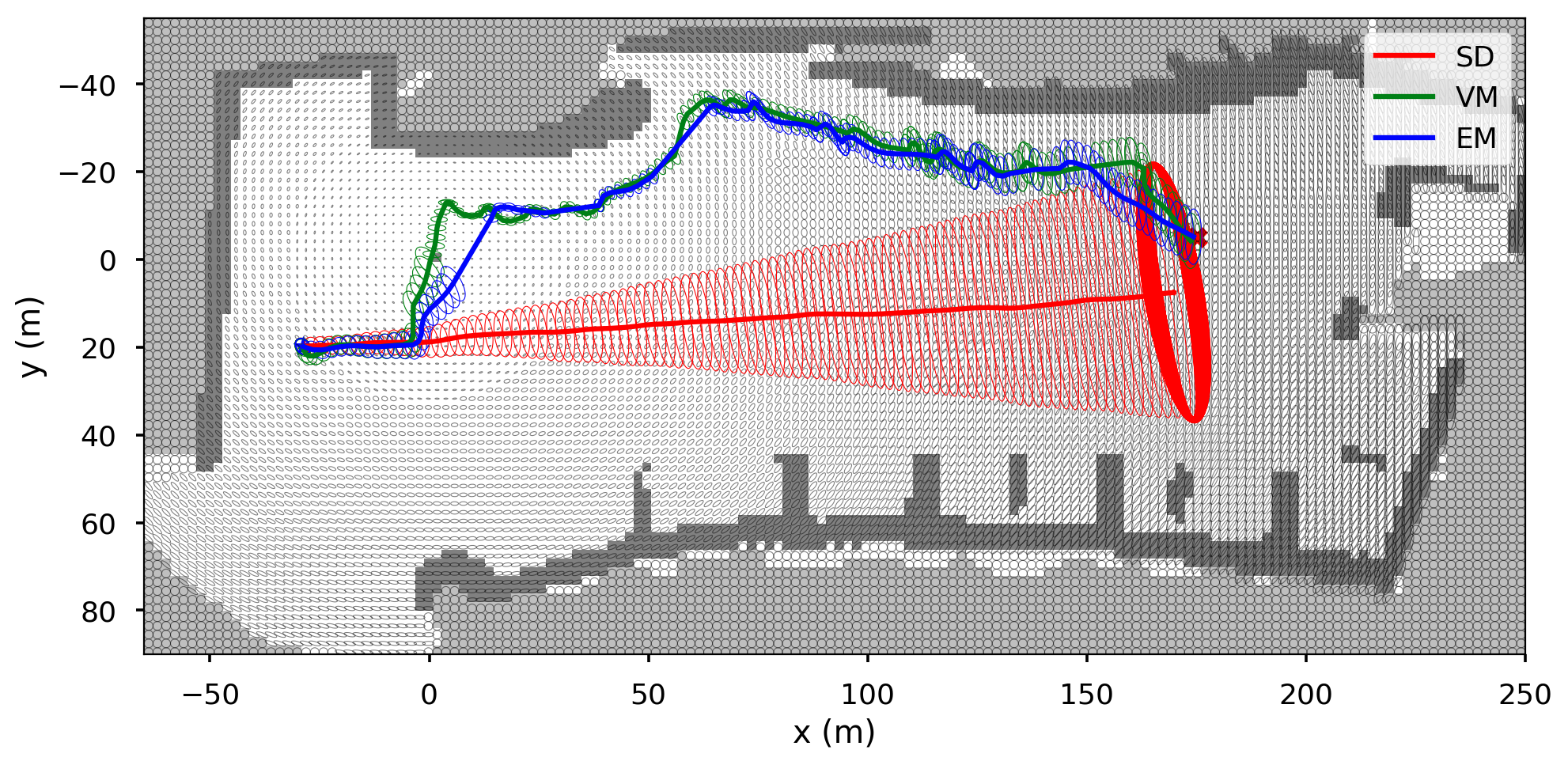}
        \label{fig:map}
    \end{subfigure}
    \caption{\textbf{Planning under uncertainty example.} \textbf{Top:} Gazebo environment and planning example. Blue star is start location, red cross shows goal location, pink arrows show candidate place-revisiting actions, and green diamond represents goal-reaching candidate action. 
    \textbf{Bottom:} Virtual map and representative resulting paths using different uncertainty handling strategies. \textit{Shortest Distance} is shown in red, \textit{Expectation Maximization} in blue, and our proposed strategy in green. The uncertainty growth from executing each path is illustrated along the path using 95\% confidence ellipses.}
    \label{fig:example}
    \vspace{-6mm}
\end{figure}

Recent 
work has focused on reducing map and robot pose uncertainty during the exploration of unknown environments, without considering navigation tasks that follow the mapping mission \cite{Palomeras2019}. Furthermore, strategies that do consider goal-reaching tasks commonly assume no prior knowledge \cite{Pairet2021} or perfect prior knowledge of the environment. However, exploration is an imperfect process, 
and the resulting maps 
are corrupted by noise, uncertainty, and incomplete information, and should be treated accordingly when used for subsequent planning queries. Planning to a specified goal 
can similarly be corrupted by noise and uncertainty. Choosing an uninformative trajectory to reach a goal can lead to substantial uncertainty growth, causing robots to crash or become "lost". Instead of discarding SLAM graph information after exploration, this information can be used to assess the uncertainty of the resulting map, informing any subsequent path planning queries. Moreover, by performing SLAM during navigation, we can adapt to small changes in the environment, manage errors from noise, and deal with incomplete information. 

In this paper, we leverage mapping and SLAM graph information from a prior exploration run and use it to create a \textit{virtual map} that will in turn be used to perform efficient planning under uncertainty, 
without the need for computationally demanding online belief propagation. Specifically, we present a planning under uncertainty scheme for underwater robots operating at a fixed depth in large-scale feature-sparse environments. Our main contributions are as follows:
\begin{itemize}
    \item To our knowledge, the first algorithmic pipeline for underwater robots that uses exploration as a pre-processing step for planning under uncertainty to specified goals. 
    \item 
    We leverage the resulting SLAM graph and occupancy map information to produce a useful costmap for computationally efficient planning under uncertainty. 
    \item Realistic simulation experiments, in which receding horizon planning is used to compare our proposed method with shortest-distance and full belief propagation planning strategies.
\end{itemize}

The rest of this work is structured as follows. First, Sec. \ref{sec:relatedworks} describes related works. Next, the problem formulation is explained in Sec. \ref{sec:problem}, followed by the proposed method in Sec. \ref{sec:method}. Subsequently, results and experiments are presented in Sec. \ref{sec:experiments}. Finally, in Sec. \ref{sec:conclusion} conclusions are drawn. 



\vspace{-2mm}
\section{Related Works}
\label{sec:relatedworks}
\vspace{-1mm}

Incorporating localization uncertainty into the planning process is important for navigation tasks including exploration and goal-reaching missions. Actively maintaining high localization accuracy while navigating was first addressed in \cite{Burgard1997}, by controlling robot motion and sensing actions. Many mobile robots, including AUVs, rely heavily on perceptual features to localize in their environment, thus perception has a substantial impact on pose uncertainty and has been studied in several works \cite{Papachristos2017, Costante2017, Zhang2018}. Moreover, in the context of SLAM, revisiting previously seen features is necessary to trigger loop closures and lower uncertainty. Image saliency was used in underwater active SLAM \cite{Suresh2020} to identify uncertainty reducing actions. Some methods \cite{Maurovic2018, Andrade2022} attempt to reduce uncertainty by performing rotational actions to observe visual features while following a path. However, only rotational movements are not sufficient to localize when features are out of sensing range.

Therefore, balancing the trade-off between reducing uncertainty and completing an exploration or goal-reaching task is an important part of planning under uncertainty strategies. To limit deviations from the main path, \cite{Suresh2020} and \cite{Andrade2022} only compute and perform uncertainty reducing actions when the uncertainty of the robot pose is above a certain threshold. Most active SLAM motion planning approaches employ three stages: goal identification, utility computation, and action selection \cite{Placed2023}. Furthermore, the receding horizon path planning strategy, in which robots execute only a small portion of each plan prior to replanning, has been used to solve the active SLAM problem in numerous works \cite{Zhang2018, Suresh2020}, where the continuous feedback of mapping the environment helps attenuate errors in tracking and perception. 

Furthermore, planning under uncertainty has been tackled using Partially Observable Markov Decision Process (POMDP)s \cite{Kaelbrling1998}. However, POMDP approaches grow exponentially in complexity as the number of actions and observations increases. To reduce computational burden various strategies have been proposed. The Belief Roadmap (BRM) \cite{Prentice2009} made belief space planning more efficient by factoring the covariance matrix and combining multiple EKF update steps. Still, it results in signiﬁcant computation time for running on a complex graph in practical problems. Rapidly-exploring Random Belief Trees (RRBTs) \cite{Bry2011} extend the fast computing Rapidly-exploring Random Tree (RRT)* \cite{Karaman2011} framework to take pose uncertainty into account. 

In the above methods, performing perception-related belief propagation is crucial, but often the computational bottleneck. Hence, another body of work has focused on developing dedicated representations of the environment and the uncertainty for efficient computation. A coarse-resolution map for rapidly evaluating the information gain of planned actions was used in \cite{Nelson2018}. Similarly, working with a reduced local map instead of a full map was done in \cite{Andrade2021}. To avoid 
online belief propagation, \cite{Roy1999} pre-computes and saves the information content of the environment in a 2D grid which can be used during the robot's mission. 
\cite{Liu2012} proposes a localizability matrix over a probability grid map which includes localizability index and direction. Yet, this approach only considers active localization and does not perform path planning. In \cite{Inoue2016}, the most feature-rich path for a rover is calculated by analyzing the soil map of Mars, assuming a constant covariance accumulation rate in feature-rich and feature-sparse areas. A technique that turns the information from publicly available maps into a localizability map is presented in \cite{Vysotska2017}. This approach uses graph SLAM, but does not incorporate path planning.

Expectation-Maximization (EM) exploration \cite{Jinkin2020, Jinkun2019, Jinkun2022} presents the concept of a \textit{virtual map}, in which every map cell's uncertainty is represented by a multivariate Gaussian, and the map uncertainty resulting from future actions can be predicted. 
In our present work, we do not focus on exploration, but on the subsequent planning to specified goal states after exploration has been performed. We adapt the virtual map presented in \cite{Jinkin2020}, and use it as an uncertainty cost map for computationally efficient planning, avoiding the need for online belief propagation. Additionally, we use a receding horizon path planning strategy that manages a goal-reaching and place-revisiting trade-off. In the sections below, we will describe our proposed planning under uncertainty framework and its application to underwater navigation with sonar, demonstrating that it achieves far lower localization uncertainty and error than a standard shortest-distance approach, and that the errors obtained are comparable to that of full-fledged EM belief propagation.

\vspace{-2mm}
\section{\color{black} Background and \color{black}{Problem Formulation}
}
\label{sec:problem}
\vspace{-1mm}

\subsection{Simultaneous Localization And Mapping}
We use an incremental 
graph-based pose SLAM approach. Thus, the entire robot trajectory is repeatedly estimated, allowing the robot pose history and map to be corrected throughout the mission. The motion model is defined by
\begin{equation}
    \label{eq:motion model}
    \mathbf{x}_i = f_i(\mathbf{x}_{i-1}, \mathbf{u}_i) + \mathbf{w}_i , \quad \mathbf{w}_i \sim \mathcal N( 0, \mathbf{Q}_i),
\end{equation}
where $\mathbf{x}_i = [x_i, y_i, \theta_i]$ is the state vector \color{black} describing position and orientation in $\mathbf{SE(2)}$\color{black}, $\mathbf{u}_i $ is the control input vector, and $\mathbf{w}_i$ is additive process noise with covariance matrix $\mathbf{Q}_i$. The sensor model is given by 
\begin{equation}
    \label{eq:sensor model}
    \mathbf{z}_{k} = h_{k}(\mathbf{x}_{ik}) + \mathbf{q}_{k}, \quad \mathbf{q}_{k} \sim \mathcal N( 0, \mathbf{R}_{k}),
\end{equation}
where we obtain measurements $\mathbf{z}_{k}$ at pose $\mathbf{x}_i$, corrupted by additive sensor noise $\mathbf{q}_{k}$ with covariance matrix $\mathbf{R}_{k}$. 

Let $\mathcal{X} = \{\mathbf{x}^{T}_{i=0}\}$  be the set of poses from time 0 to time $T$. $\mathcal{C}$ contains all constraints between robot poses. For each pair of poses defining a constraint $\langle i,j \rangle \in \mathcal{C}$, the error $\mathbf{e}_{ij}$ between observed transformation $\mathbf{z}_{ij}$ and predicted transformation  $\hat{\mathbf{z}}_{ij}$ is defined as:
\begin{equation}
    \label{eq:error}
    \mathbf{e}_{ij}(\mathbf{x}_i, \mathbf{x}_j) = \mathbf{z}_{ij} - \hat{\mathbf{z}}_{ij}(\mathbf{x}_{i}, \mathbf{x}_j),
\end{equation}
\begin{equation}
    \label{eq:error1}
    \hat{\mathbf{z}}_{ij}(\mathbf{x}_{i}, \mathbf{x}_j) = \mathbf{x}_i ^ \top \mathbf{x}_j.
\end{equation}
The SLAM problem can be represented as a nonlinear least squares problem \cite{Grisetti2010}: 
\begin{equation}
    \label{eq:error2}
    \mathcal{X}^* = {\underset{{\mathbf{X}}}{\arg\min}} \sum_{\langle i,j \rangle \in \mathcal{C}} \mathbf{F}_{ij},
    \qquad
    \mathbf{F}_{ij} = \mathbf{e}_{ij}^ \top \mathbf{\Omega}_{ij} \mathbf{e}_{ij},
\end{equation}
where $\mathbf{F}_{ij}$ denotes the negative log-likelihood function of one constraint between $\mathbf{x}_i$ and $\mathbf{x}_j$, and $\mathbf{\Omega}_{ij}$ is the information matrix. The SLAM system aims to find a set of robot poses which minimize the total observation error. 

\vspace{-1mm}
\subsection{Virtual Map}


Unlike occupancy grid maps, the virtual map \cite{Jinkun2022} encodes map cell uncertainty (represented by a multivariate Gaussian) instead of occupancy probability. Let the virtual map $\mathcal{V}=\{\mathbf{v}_i\}$ be the set of 
possible observations located at each map cell $\mathbf{v}_i \in  	\mathbb{R}^2$, distributed as $\mathcal{N}(\mathbf{v}_i, \Sigma_{\mathbf{v}_i})$, where $\mathbf{v}_i = [x_i, y_i]$ is the cell location and $\Sigma_{\mathbf{v}_i}$ is the corresponding covariance matrix. When initializing a virtual map, a prior with conservatively high initial uncertainty is imposed on all map cells. 


The map cell covariances are updated based on the current estimate of the trajectory $\mathcal{X}_\text{old}$ and the history of measurements $\mathcal{Z}$.  \vspace{-2mm}
\begin{equation}
    \label{eq:V}
   q(\mathcal{V}) = p(\mathcal{V|X}_{\text{old}}, \mathcal{Z}) 
\end{equation}
The maximum posterior probability estimate for the virtual map can thus be computed by
\begin{equation}
    \label{eq:V*}
   \mathcal{V}^* = \underset{\mathcal{V}}{\arg\max} \ p(\mathcal{V|X}_{\text{old}}, \mathcal{Z}).
\end{equation}
It is assumed that measurements are assigned to maximize the likelihood \vspace{-2mm}
\begin{equation} \vspace{-2mm}
    \mathcal{Z} = \underset{\mathcal{Z}}{\arg\max} \ h(\mathcal{X,V}).
\end{equation}

By measuring environmental features, we can derive the transformations between poses with overlapping observations using Iterative Closest Point (ICP). We essentially aim to minimize the closeness of measurements using ICP-based pose SLAM. For more details regarding virtual map pose covariance computation, the reader should consult \cite{Jinkun2022}. 
\vspace{-1mm}
\subsection{\color{black}Problem Statement \color{black}}

In this paper, the problem of autonomously navigating to a goal location under pose uncertainty is addressed. We adapt the virtual map, previously used for exploration, to accommodate planning under uncertainty to specified goals. We leverage SLAM information from a prior exploration run to compute uncertainty values offline and store them in the form of a virtual map $\mathcal{V}$. We then focus on finding a path to a goal that curbs the growth of uncertainty using pose SLAM and the covariance information from our virtual map. \color{black} Specifically, \color{black} the virtual map is used as a cost map during the planning stage to estimate future uncertainty, \color{black} and the SLAM graph is maintained and updated to help mitigate dead reckoning errors and to update our occupancy map used for planning, as the prior map is imperfect.\color{black}

\vspace{-2mm}
\section{Methodology} \vspace{-1mm}
\label{sec:method}
\begin{figure} 
    \centering
    \includegraphics[width=0.35 \textwidth]{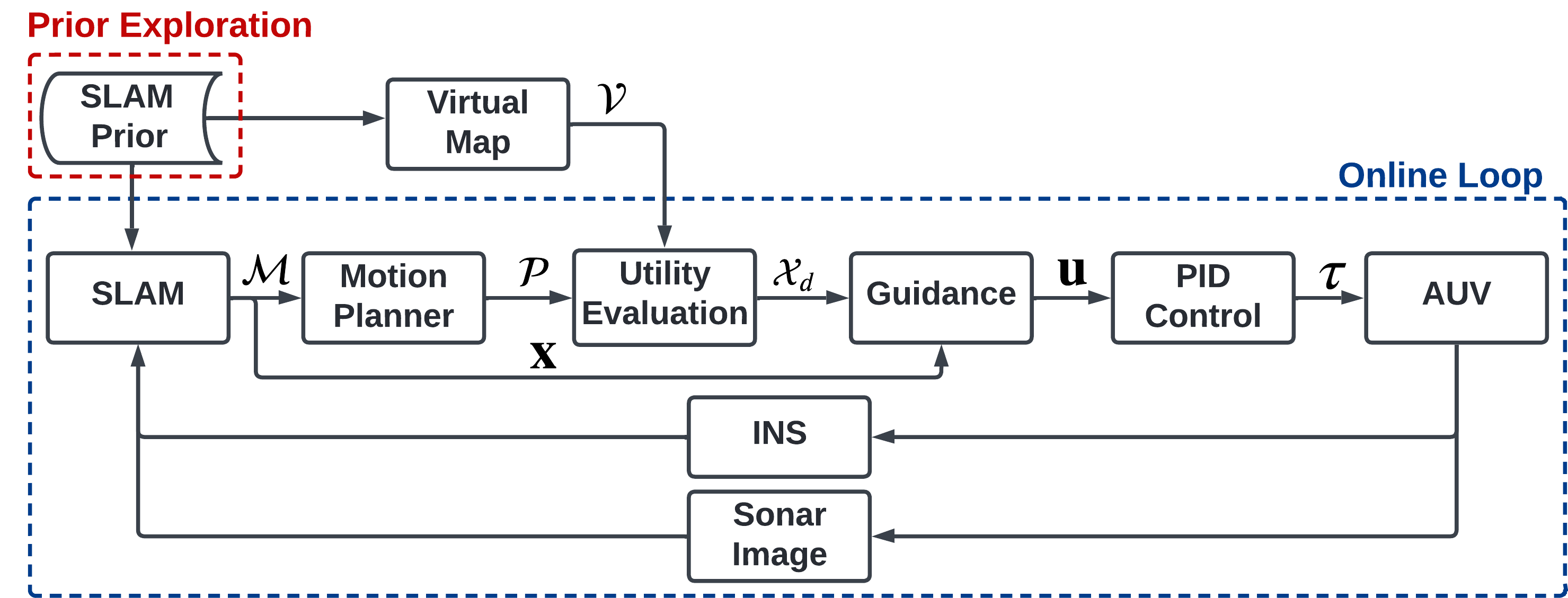}
    \caption{\textbf{System overview}: SLAM information from a prior exploration run is used to initialize the current SLAM instance and build a virtual map offline. The motion planner takes the accompanying occupancy map 
    and searches for candidate paths. The utility of each candidate path is assessed using the virtual map. The chosen path and the current robot pose are then passed to the guidance module, which provides the desired velocity vector to follow the desired path. The low-level PID controller takes the desired velocity values and defines the required thruster forces.}
    \label{fig:overview}
    \vspace{-5mm}
\end{figure}
In this section, we explain our proposed framework, the system overview can be observed in Fig. \ref{fig:overview}. First, SLAM information from a prior exploration run is taken and used to initialize our current SLAM instance and build our virtual map $\mathcal{V}$. The motion planner takes the occupancy map $\mathcal{M}$ from SLAM and searches for candidate paths $\mathcal{P} = \{\mathcal{X}_c\}$. Then, the utility of each candidate path is assessed using the virtual map. The chosen path $\mathcal{X}_d$ and the current robot pose $\mathbf{x}$ are then passed to the pure pursuit guidance module, which in turn provides the appropriate desired velocity vector $\mathbf{u}$ to follow the desired path. The low-level PID controller takes the desired velocity values and defines 
thruster forces $\mathbf \tau$. 

\vspace{-2mm}
\subsection{Sonar SLAM}
\label{Sonar SLAM}
To estimate the location of the robot, we use the same graph-based pose SLAM framework as in \cite{Jinkun2022}. The sensory inputs include an Inertial Measurement Unit (IMU), a Doppler Velocity Log (DVL), and  a sonar image. The sonar image is first processed using Constant False Alarm Rate (CFAR) detection \cite{Richards2005} to identify returns from the surrounding environment. All CFAR-extracted features in the image are converted to Cartesian coordinates.
 The SLAM factor graph is denoted as
\begin{flalign*} \vspace{-1mm}
\mathbf f(\boldsymbol \Theta) = \mathbf  f^{\text{0}}(\boldsymbol \Theta_0) & \prod_i \mathbf f^{\text{O}}_{i}(\boldsymbol \Theta_i)  \prod_j \mathbf f^{\text{LC}}_j(\boldsymbol \Theta_j), \vspace{-1mm}
\end{flalign*}
where variables $\boldsymbol{\Theta}$ contain 3-DOF robot poses \color{black} and $\mathbf  f^{\text{0}}$ denotes the initial factor graph. \color{black} The odometry factor $\mathbf{f}^\text{O}$ defines the relative motion between two consecutive poses from persistent odometry measurements obtained by DVL/IMU dead reckoning. Loop closure factors  $\mathbf{f}^\text{LC}$ are obtained by identifying non-sequential scan matches, applying  Iterative Closest Point (ICP) \cite{Besl1992} between the current sonar keyframe and frames inside a given search radius. ICP is initialized using a DVL/IMU dead reckoning pose estimate and further optimized using consensus set maximization \cite{Fischler1987}, helping ICP to avoid local minima. Loop closure outliers are rejected by evaluating point cloud overlap and applying Pairwise Consistent Measurement Set Maximization (PCM) \cite{Mangelson2018}. At the back end, we employ the GTSAM \cite{Dellaert2012} implementation of iSAM2 \cite{Kaess2012}, which is merged with dead reckoning to provide a high-frequency state estimate. \color{black} Occupancy map $\mathcal M $ is updated and used for path planning. \color{black}



\begin{algorithm}[tb]
\caption{Place-revisiting actions}
\label{alg:place-revisiting}
\begin{algorithmic}
\Require {Occupied cells $\mathcal O$, number of place-revisiting actions $N_\text{pr}$, revisiting radius $r_\text{pr}$, minimum separation distance $d_{\text{pr}}$, shortest path to goal $\mathcal{X}_\text{sd}$, \color{black} number of clusters $k$\color{black}}
\Ensure $\mathcal A_{\text{pr}}, \mathcal R \leftarrow \emptyset$
\State $\mathcal M = \{\mathbf m_{1, ..., k}\} \leftarrow \text{ComputeKMeans}(\mathcal O, k)$
\For {$i \gets 1$ to $k$}
\State $\mathcal C \leftarrow \{(x_i + r_\text{pr}\cos(\theta), y_i + 
r_\text{pr}\sin(\theta), \color{black} \theta \color{black}) \ | \ \theta \in [0, 2\pi)\}$
\State $\mathcal R \leftarrow \mathcal R \cup \{\mathcal C\}$
\EndFor
\While {$|\mathcal A_{\text{pr}}| < N_\text{pr}$}
\State $\mathbf a* \leftarrow\arg \min _{\mathbf r_i \in \mathcal R} ||\mathbf r_{i}-\mathcal{X}_\text{sd}||^2$

\State $d^* \leftarrow \min_{\mathbf g_i \in \mathcal A_{\text{pr}}} |\mathbf a^* - \mathbf g_i|$
\If {$d^* \le d_\text{pr}$}
\State $\mathcal A_{\text{pr}} \leftarrow \mathcal A_{\text{pr}} \cup \mathbf a^*$
\EndIf
\State $\mathcal R \leftarrow \mathcal R  \setminus \{\mathbf a^*\}$
\EndWhile
\State \Return {Candidate place-revisiting actions $\mathcal A_{\text{pr}}$}
\end{algorithmic}
\vspace{-5mm}
\end{algorithm}

\vspace{-10mm}
\subsection{Motion Planning}
\label{Planner}
Global trajectory optimization in a large-scale planning problem is computationally expensive, and may require replanning as the trajectory is executed. 
Therefore, we apply a receding horizon path planner, which also performs trajectory tracking and incorporates perceptual feedback. 
Inspired by many active SLAM motion planning approaches \cite{Placed2023}, we employ three stages: goal identification, utility computation, and action selection. Action selection and path planning are treated as tightly coupled problems. We consider two different action types, one type for lowering uncertainty (place-revisiting), and another for reaching the mission goal (shortest-distance). \textit{Place-revisiting} actions look for poses that can lower uncertainty by revisiting locations the robot has previously observed, triggering SLAM loop closures. \textit{Shortest-distance} actions represent locations that lay on the shortest feasible path to the goal. Fig. \ref{fig:example}a depicts place-revisiting actions as pink arrows and the shortest-distance action as a green diamond.

First, a finely discretized roadmap is generated at the beginning of the mission. The boundaries of the roadmap are defined, and edges are pruned out if obstacles are discovered. The A* algorithm \cite{Hart1968} is used to find the shortest path from the start position to the final mission goal without colliding with known obstacles. Afterward, the shortest-distance action set $\mathcal{A}_\text{sd}$ is defined by choosing targets at user-defined distances $d_\text{sd}$ from the current position along the identified shortest distance path. For safety, the distance to the waypoints should be less or equal to the robot sensing range to avoid collision with possible new obstacles and refrain from unnecessary replanning. Likewise, traversing large featureless distances should be prevented to avoid large uncertainty growth. 

The place-revisiting action set $\mathcal{A}_\text{pr}$ is defined following Algorithm \ref{alg:place-revisiting}. The target locations are chosen by identifying action candidates that are close to the shortest path while observing occupied cells. These poses have the potential of triggering SLAM loop closures while avoiding large deviations from the desired path. First, occupied cells in the occupancy map are identified and clustered using the k-means algorithm. Afterward, candidate goals are sampled along the boundary of a circle centered at each cluster origin with a radius $r_\text{pr}$ that is user-specified. We then choose the candidates that are closest to the shortest path. Finally, the A* search algorithm is used to find paths from the current position to the final goal, each path passing through one place-revisiting waypoint. 

From the candidate paths $\mathcal{P}$ identified, one is chosen using the utility function defined in Sec. \ref{sec:utility}. Once the desired path $\mathcal{X}_\text{d}$ is identified, 
and the vehicle arrives at the place-revisiting or shortest-distance target location, the planning strategy is repeated until the robot arrives at its final goal. 

\subsection{Utility evaluation}
\label{sec:utility}
We define the utility function used to choose the desired path $\mathcal{X}_\text{d} = \arg \max_{X \in \mathcal P} U(X_{T:T+N})$ as follows:
\begin{equation}
\label{eq:Utility}
    \begin{aligned}
        U(X_{T:T+N}) = 
        &- \sum^{n}_{i=1} \log \det(\Sigma_{\mathbf{v}_i}) - \alpha d(X_{T:T+N}).
    \end{aligned}
\end{equation} 
The uncertainty metric used here is the sum of the log-determinant of the covariance matrices of the virtual map cells $\Sigma_{\mathbf{v}_i}$ that will be encountered by the chosen path. The entire path from the current position to the target position is considered. In addition to the uncertainty term, a distance cost weighted by  $\alpha$ is added to encourage a trade-off between traveling and uncertainty reduction. A path will be discarded and recomputed if it is found to be blocked by an obstacle. 

\subsection{Virtual Map Adaptation}

In contrast to the original implementation of virtual maps, we compute virtual map values only once at the start of a planning run. First, the SLAM graph and occupancy map information from a prior SLAM exploration run are loaded. Then, the past trajectory information $\mathcal{X}_\text{old}$ is used to estimate virtual map covariances $\Sigma_{\mathbf{v}_i}$. 
Even though we can use the same map resolution for path planning and uncertainty estimation, it was shown in \cite{Jinkin2020} that a lower resolution virtual map provides similar performance but permits faster belief propagation. In the interest of computational efficiency and fair experimental comparisons, we discretize the virtual map to be as coarse as possible, while still capturing the essential geometric features of the robot's workspace. 

\section{Experiments and Results}
\label{sec:experiments}

\subsection{Experiment Setup}

To validate the proposed planning under uncertainty framework, we present experimental results from a high-fidelity AUV simulation operating at a fixed depth in different environments. The simulation setup ran on a Lambda Workstation with an AMD 3970X 32-core processing unit and 256GB RAM, running Ubuntu 18.04, ROS Melodic, and Gazebo (our planning implementation is serialized and every trial used only a single core). The vehicle used in the simulations is the RexROV2 from UUV Simulator \cite{UUVSim}. The simulated sonar operates at 5 Hz, has 30 meter range, and has a horizontal aperture of $\theta = [-65^{\circ}, 65^{\circ}]$. 
Zero-mean Gaussian noise is added to range and bearing sonar measurements: $\sigma_r = 0.2$ m, $\sigma_\theta = 0.02$ rad. Linear velocity values are obtained from a simulated Doppler Velocity Log (DVL) and angular measurements are obtained from a simulated Inertial Measurement Unit (IMU) at 50 Hz and 7 Hz respectively. Likewise, zero-mean Gaussian noise is added with $\sigma_x = \sigma_y = 0.1$ m/s and $\sigma_\theta = 0.015$ rad.

\begin{figure}
\centering
\begin{subfigure}{0.3\textwidth}
    \includegraphics[width=\textwidth]{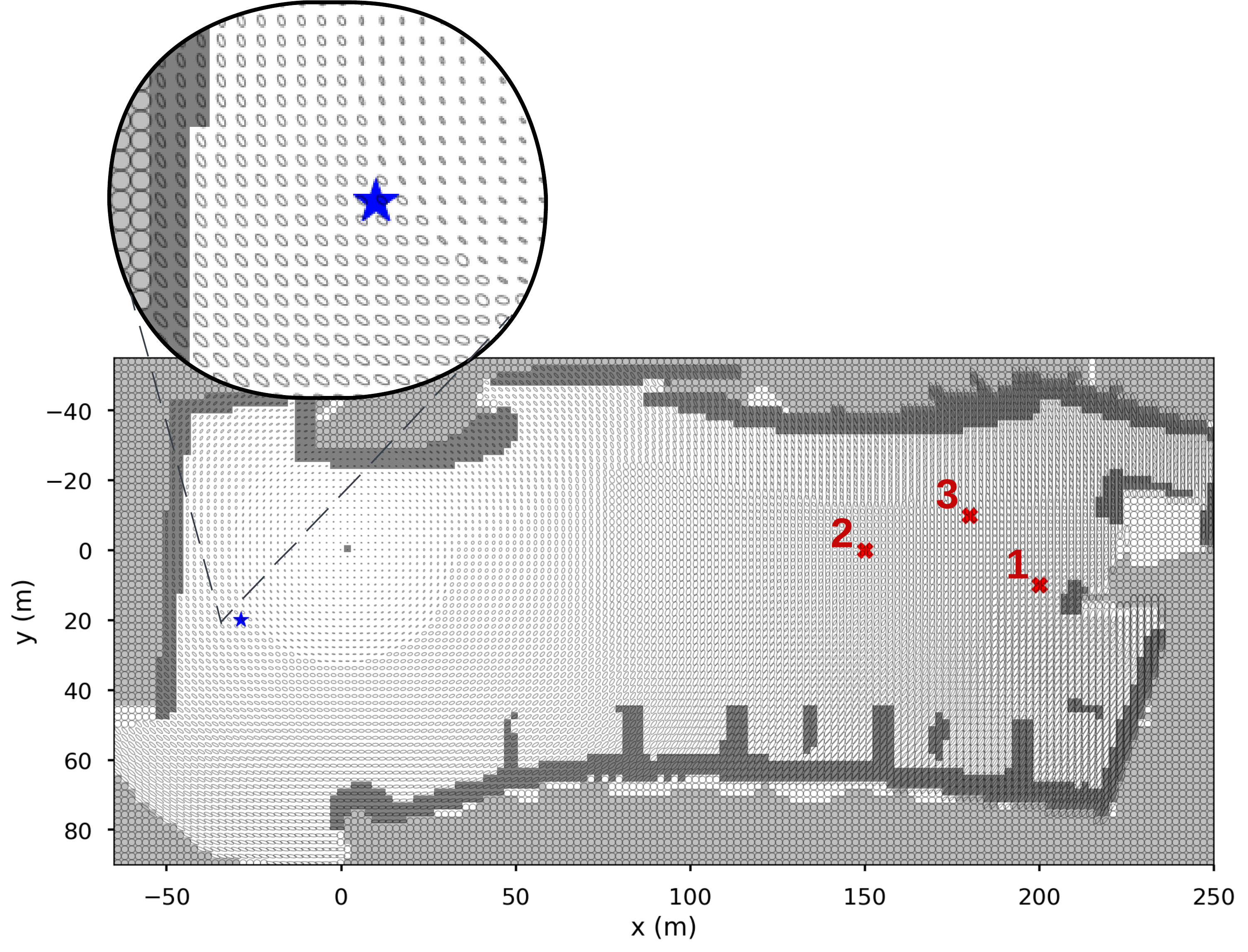}
    \caption{Marina environment.}
    \label{fig:PL_VM}
\end{subfigure}
\hspace{-5mm}
\begin{subfigure}{0.20\textwidth}
    \includegraphics[width=\textwidth]{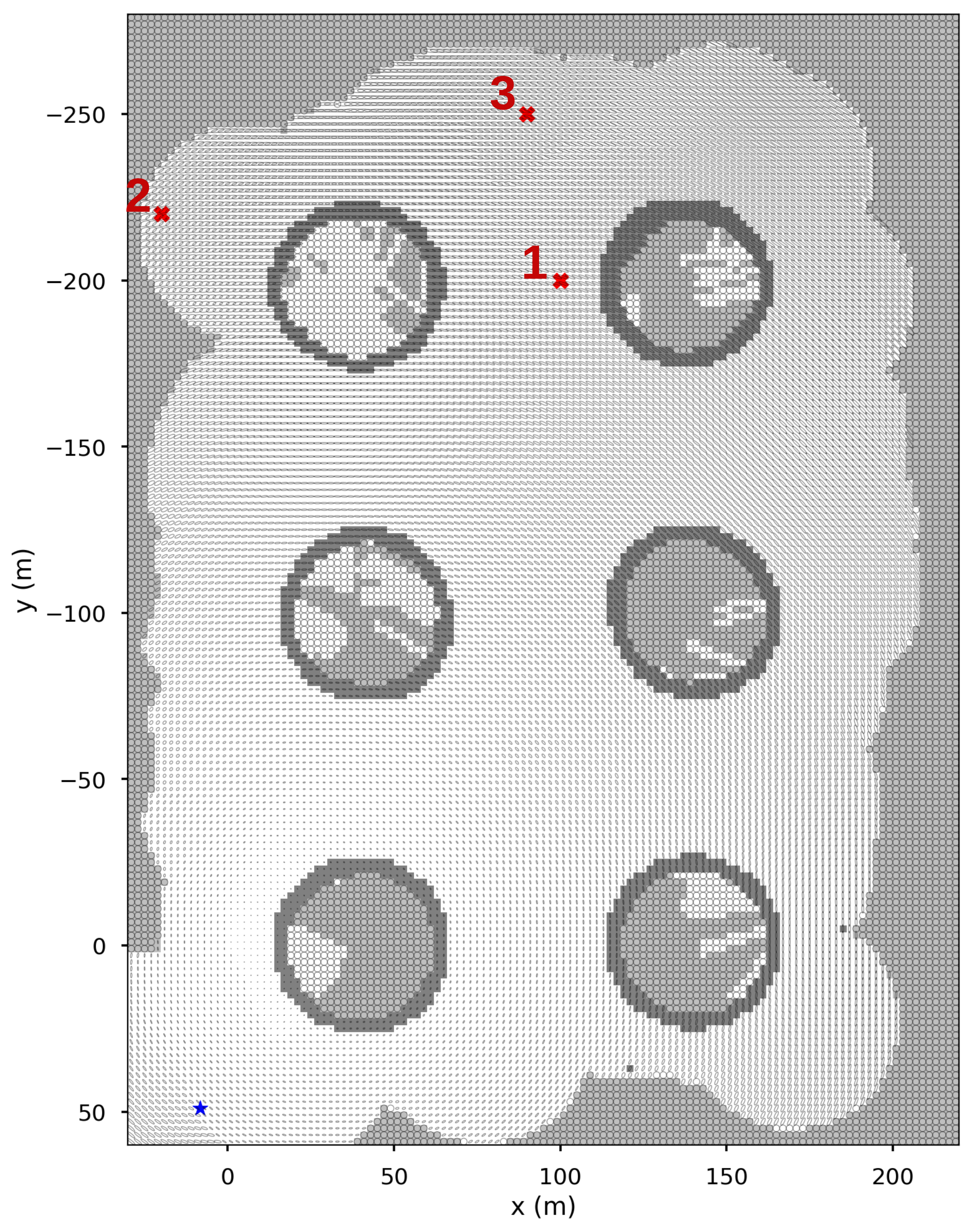}
    \caption{Fish farm environment.}
    \label{fig:FF_VM}
\end{subfigure}
\hfill
\begin{subfigure}{0.5\textwidth}
    \includegraphics[width=\textwidth]{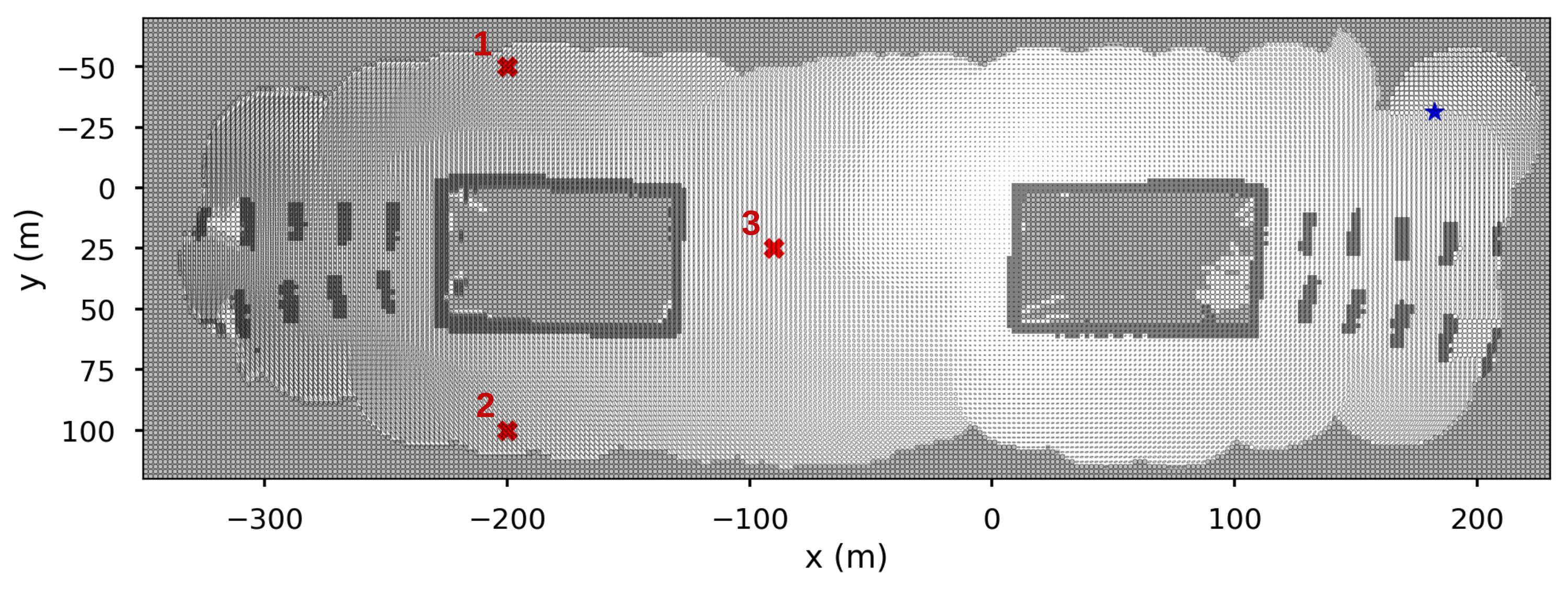}
    \caption{Bridge-tunnel environment.}
    \label{fig:C_VM}
\end{subfigure}    
\caption{\textbf{Virtual maps for all three simulation environments}: including planning start (blue star) and all goal positions (numbered red x's).}
\label{fig:VM}
\vspace{-5mm}
\end{figure}

\begin{figure}
\centering
\begin{subfigure}{0.5\textwidth}
    \includegraphics[width=0.5\textwidth]{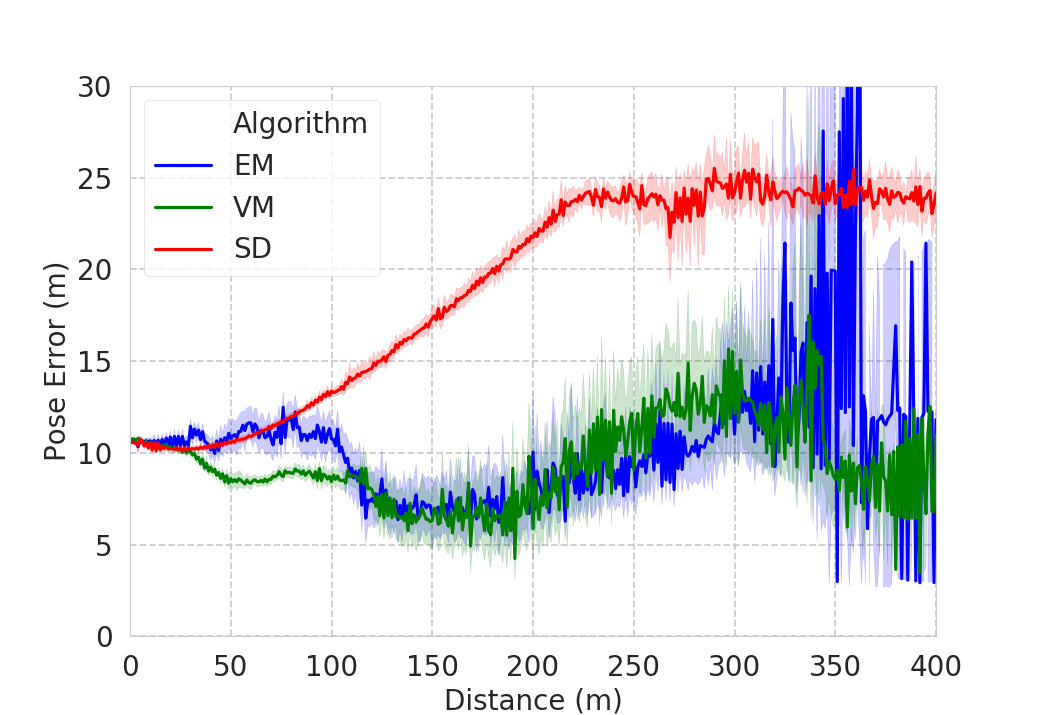}
    \hspace{-5mm}
     \includegraphics[width=0.5\textwidth]{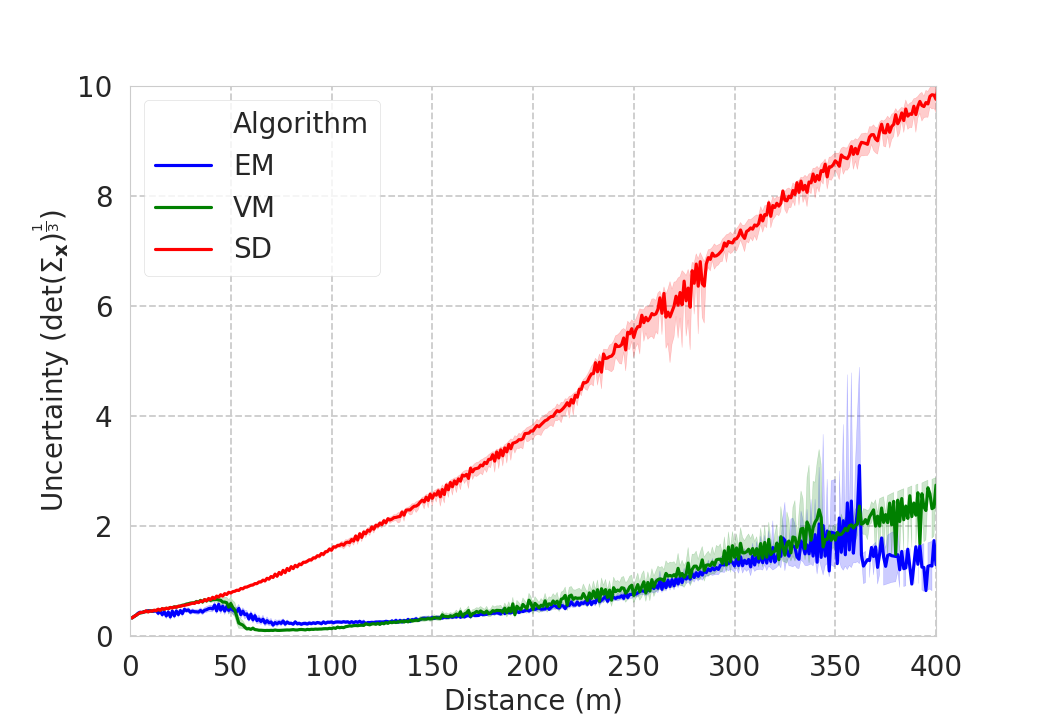}
    \label{fig:PL1}
\end{subfigure}
\hspace{-5mm}
\begin{subfigure}{0.5\textwidth}
    \includegraphics[width=0.5\textwidth]{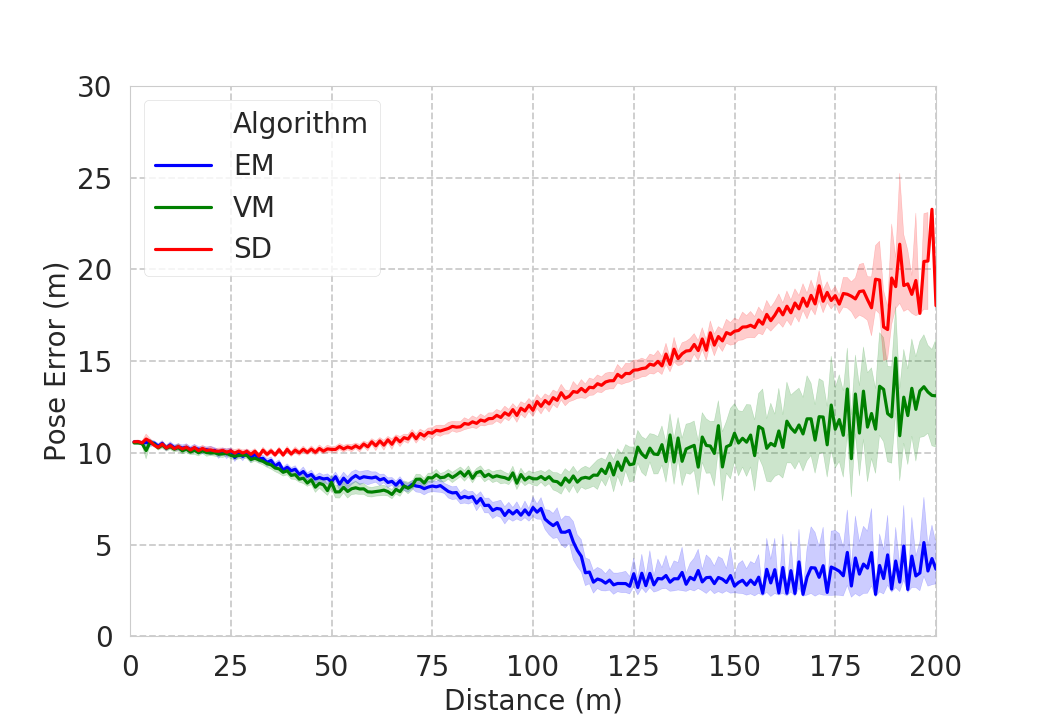}
    \hspace{-5mm}
    \includegraphics[width=0.5\textwidth]{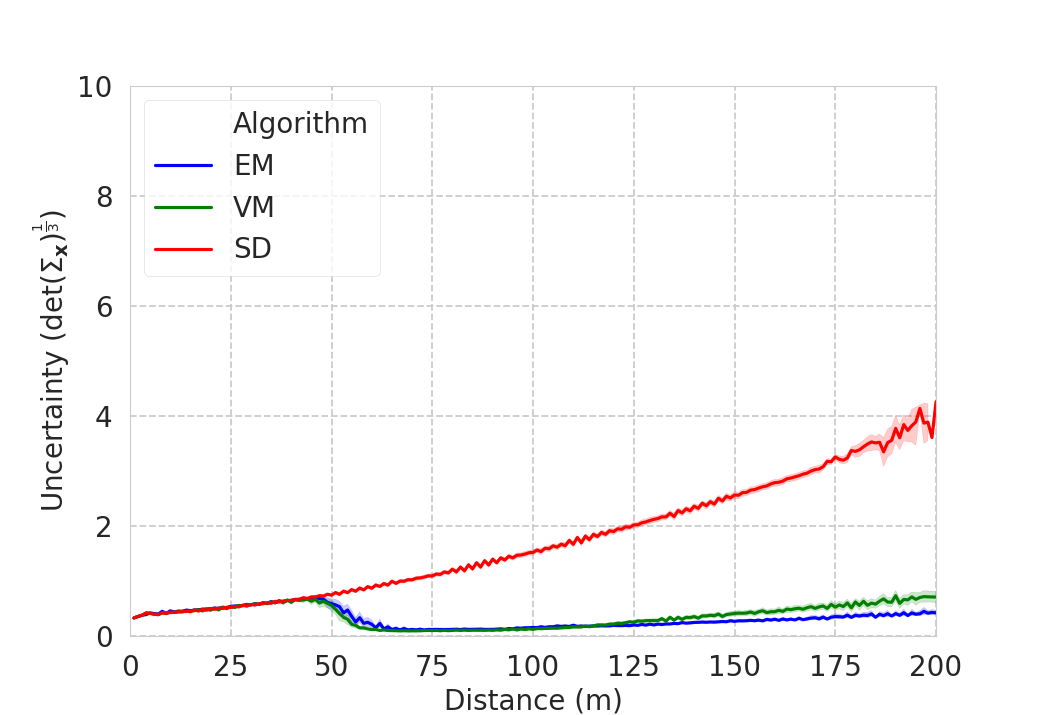}
    \label{fig:PL2}
\end{subfigure}   
\hspace{-5mm}
\begin{subfigure}{0.5\textwidth}
    \includegraphics[width=0.5\textwidth]{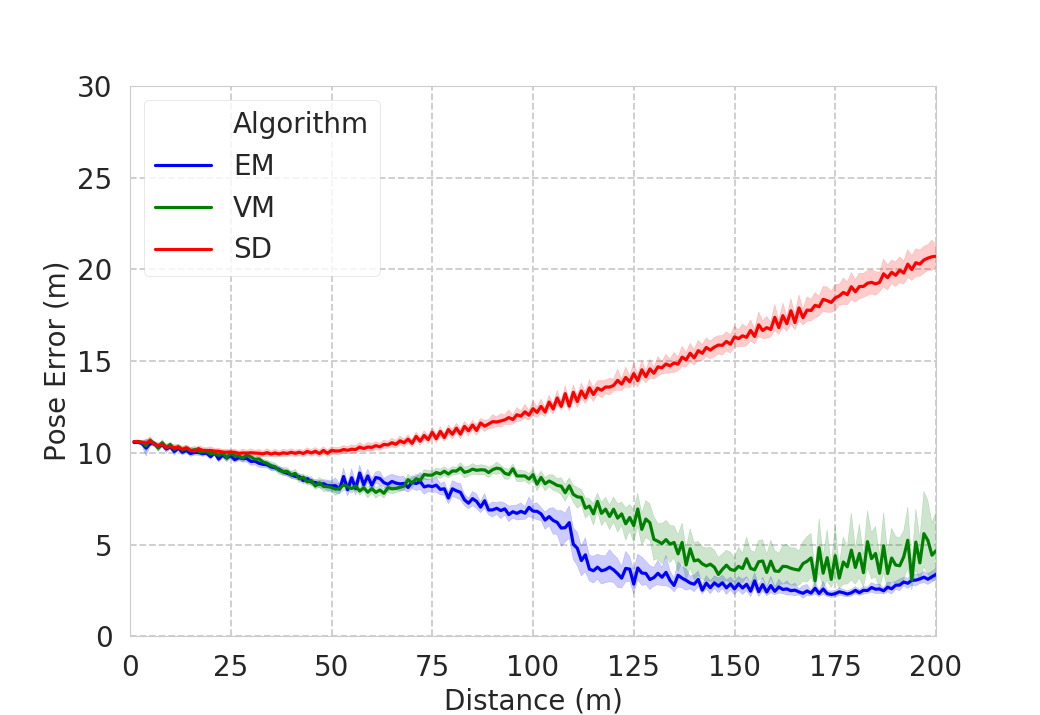}
    \hspace{-5mm}
    \includegraphics[width=0.5\textwidth]{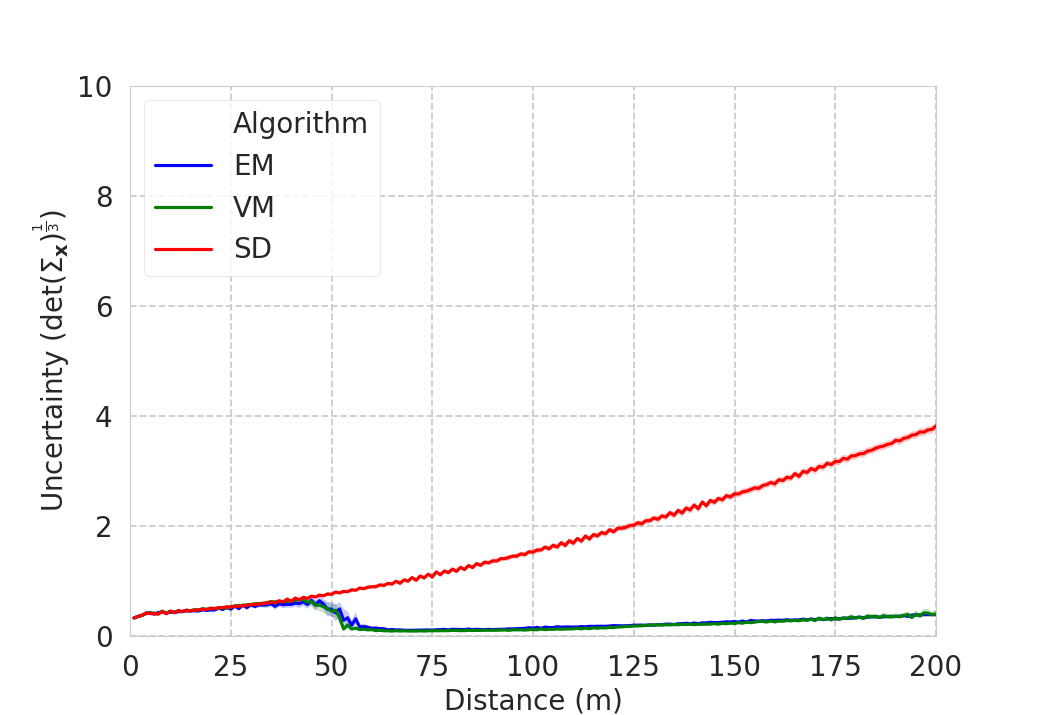}
    \label{fig:PL3}
\end{subfigure} 
\caption{\textbf{Marina results:} goals 1, 2, 3 shown from top to bottom.}
\label{fig:PL}
\vspace{-5mm}
\end{figure}

Three environments were designed to simulate realistic large-scale scenarios where an AUV might operate. The first environment, a marina, was inspired by Penn's Landing in Philadelphia, the second scenario is an offshore fish farm containing six cylindrical fish pens, and the third scenario is a bridge-tunnel system, inspired by the Chesapeake Bay bridge-tunnel.  
As a preprocessing step, each environment was fully explored while performing sonar SLAM as presented in Sec. \ref{Sonar SLAM} \cite{Jinkun2022}.
Our AUV was teleoperated inside its simulated environment to efficiently complete the exploration task, during which a globally deformable occupancy grid map was constructed based on local submaps \cite{Ho2018}. This representation anchors a local submap to each SLAM pose keyframe, allowing for efficient map recomputation when a trajectory segment is changed. 
The trajectories used for each exploration can be observed in the video attachment. The SLAM graph and occupancy map information were saved and later reloaded to build the virtual map and use our planner. The occupancy grid map was downsampled from 0.2m to 2m in cell size to produce the virtual map.

Each scenario has one start location and three different goal locations. Each run consisted of the robot navigating from a start to a goal location, and was repeated for 100 trials. The results shown are the average values of all the trials executed. Additionally, we compare our planner framework with two other strategies, Shortest Distance (SD) and Expectation-Maximization (EM). The \textit{SD planner} just follows the shortest viable A* path to the goal. The \textit{EM planner} uses the same receding horizon style from Sec. \ref{Planner}, however, it employs full Expectation-Maximization belief propagation as in \cite{Jinkun2022} to choose each intermediate goal. 

For our experiments, we chose to have one shortest-distance action $N_\text{sd} = 1$ at $d_\text{sd} = 25$ m distance, and five place revisiting actions $N_\text{pr} = 5$ each iteration. The place revisiting radius was defined at $r_\text{pr} = 15$ m. The distance weight was \color{black} experimentally chosen and \color{black} set to $\alpha=3$. To evaluate our planning framework, we recorded the pose uncertainty of the current robot pose, computed as $\text{det}(\Sigma_{\mathbf x_i})^{\frac 13}$, and the pose error, computed via RMSE, across all trials as functions of distance traveled. 

\vspace{-2mm}

\subsection{Results Comparison} 
In our results, SD refers to the \textit{Shortest Distance} planner, EM refers to the \textit{Expectation-Maximization} planner, and VM refers to our proposed planner, which uses a virtual map as a costmap to approximate belief propagation. 

\subsubsection{Marina Environment}
The virtual map for the marina environment 
can be observed in Fig. \ref{fig:PL_VM}, where the blue star represents the start location, and the red crosses depict goal locations. Pose error and uncertainty results for each goal location can be observed in Fig. \ref{fig:PL}. VM and EM planners achieve a similarly low uncertainty performance, while SD uncertainty grows unbounded. Goal 3 results in Fig. \ref{fig:PL} (bottom) show VM and EM manage to reduce the initial robot pose error, while SD does not. Surprisingly, goal 1 results in Fig. \ref{fig:PL} (top) reveal VM achieves a more stable low pose error performance than EM, likely due to EM having an erroneous loop closure. Goal 2 results in Fig. \ref{fig:PL} (middle) convey that EM has the best pose error result, while VM has intermediate performance, still better than SD. Here, VM likely missed an opportunity to lower pose error by not predicting and achieving a SLAM loop closure. 

\subsubsection{Fish Farm Environment}
\begin{figure}
\centering
\begin{subfigure}{0.5\textwidth}
    \includegraphics[width=0.5\textwidth]{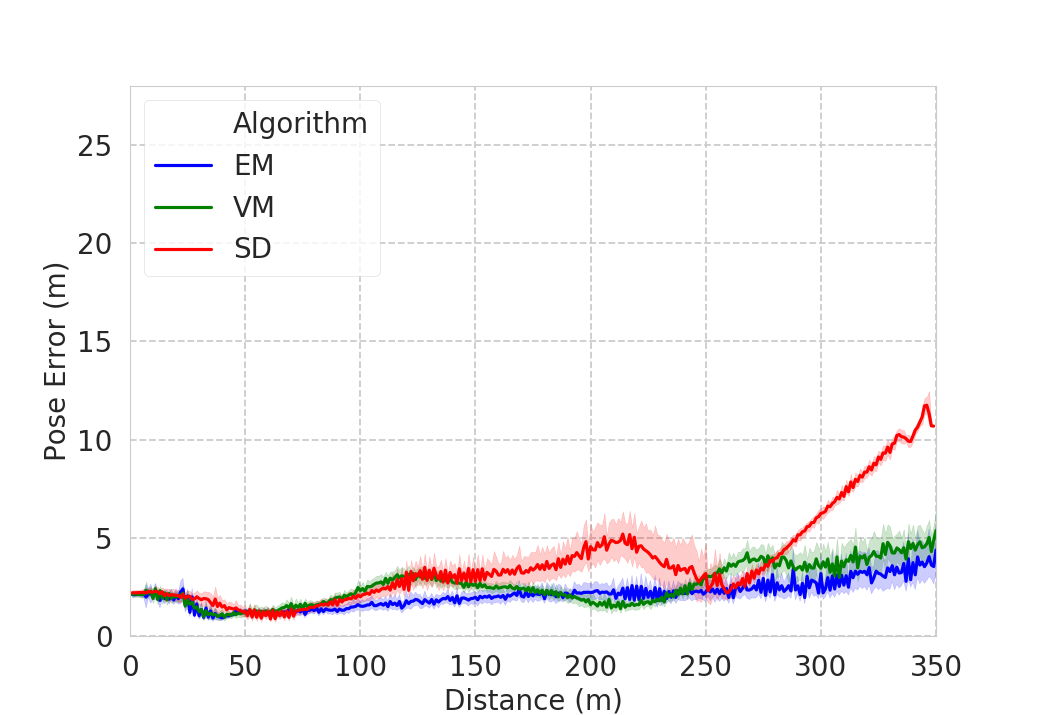}
    \hspace{-5mm}
    \includegraphics[width=0.5\textwidth]{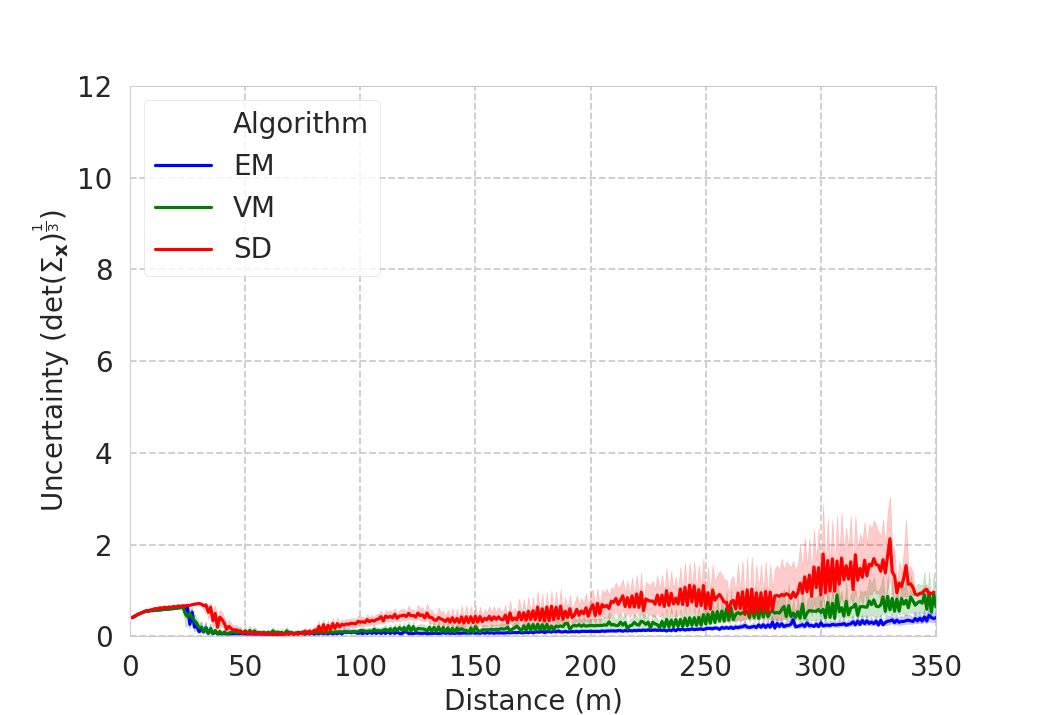}
    \label{fig:FF1}
\end{subfigure}
\hspace{-5mm}
\begin{subfigure}{0.5\textwidth}
    \includegraphics[width=0.5\textwidth]{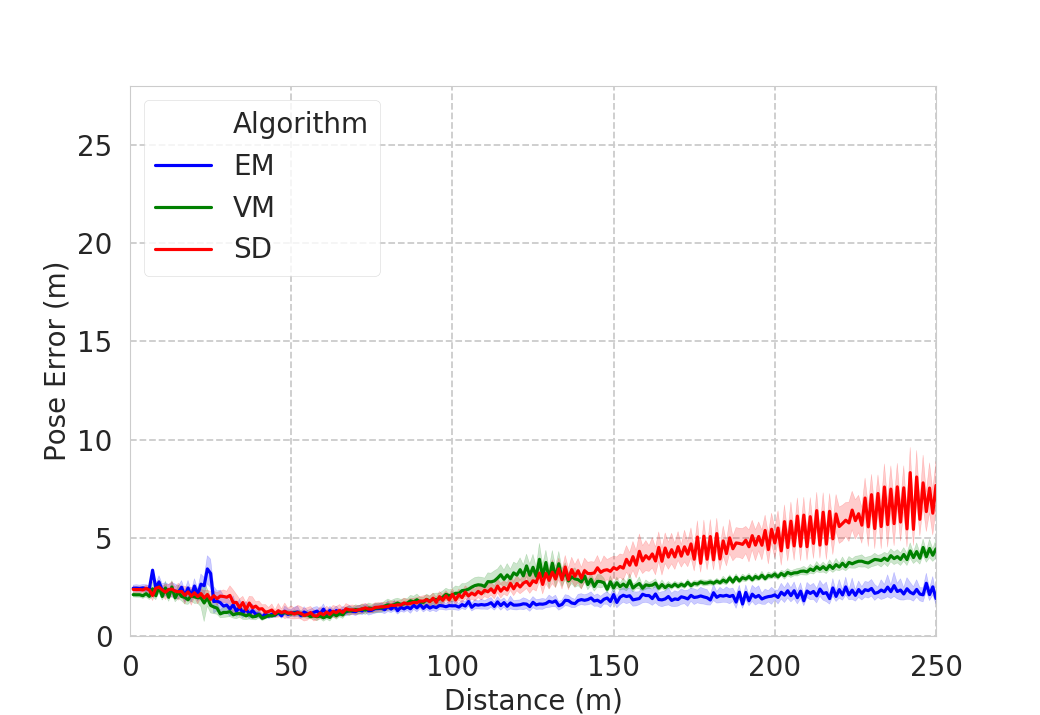}
    \hspace{-5mm}
    \includegraphics[width=0.5\textwidth]{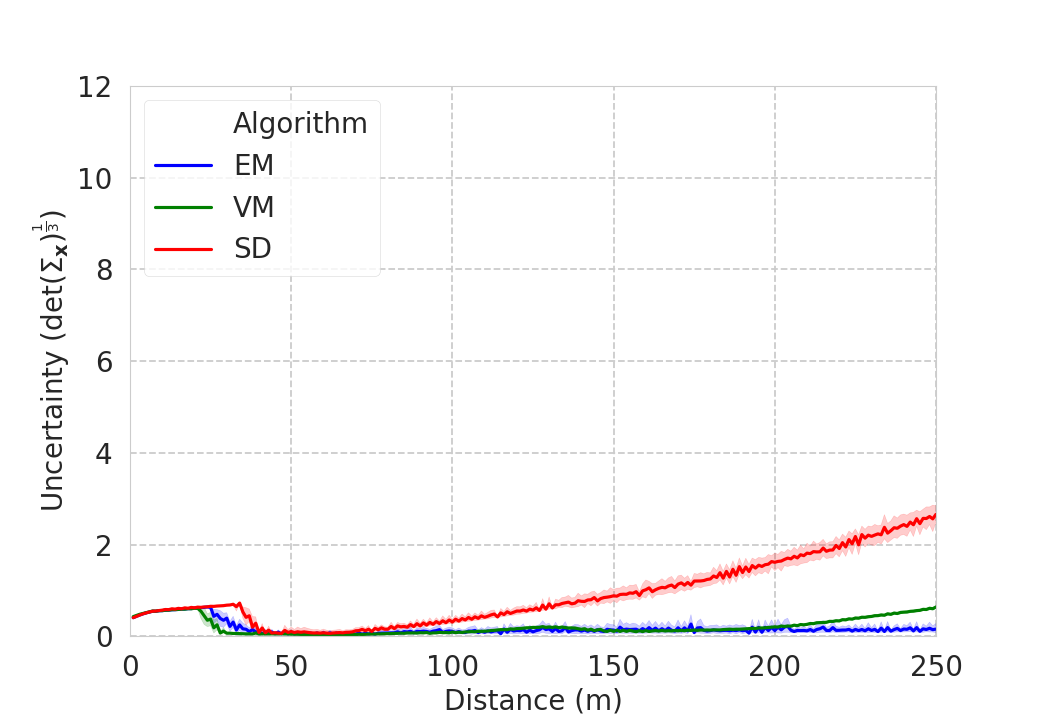}
    \label{fig:FF2}
\end{subfigure}   
\hspace{-5mm}
\begin{subfigure}{0.5\textwidth}
    \includegraphics[width=0.5\textwidth]{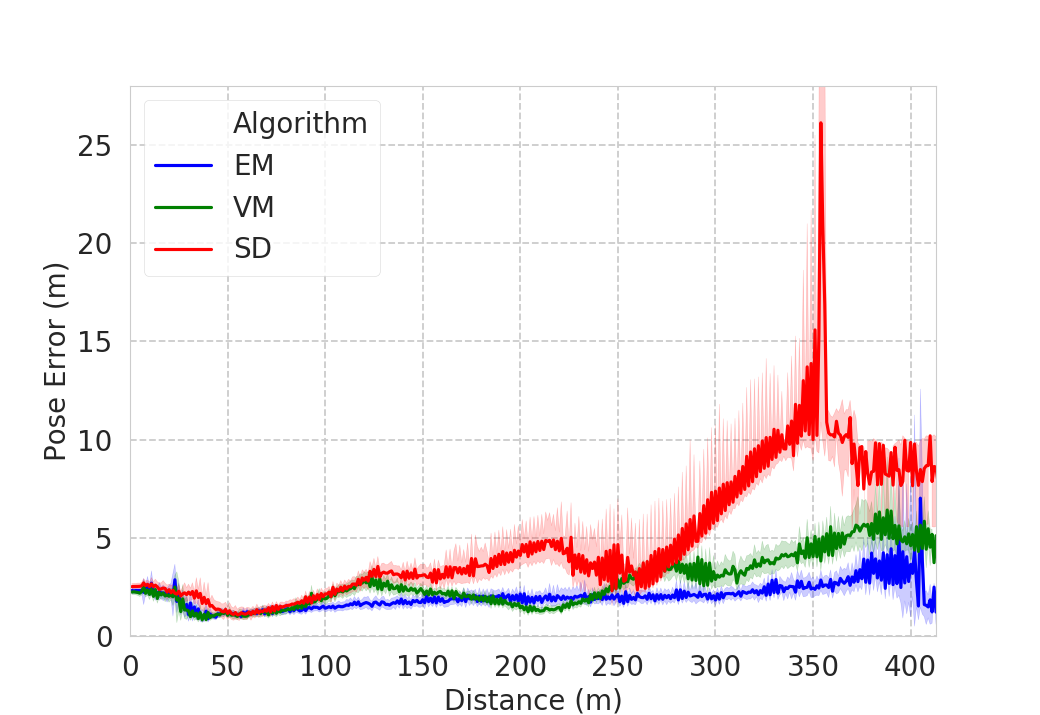}
    \hspace{-5mm}
    \includegraphics[width=0.5\textwidth]{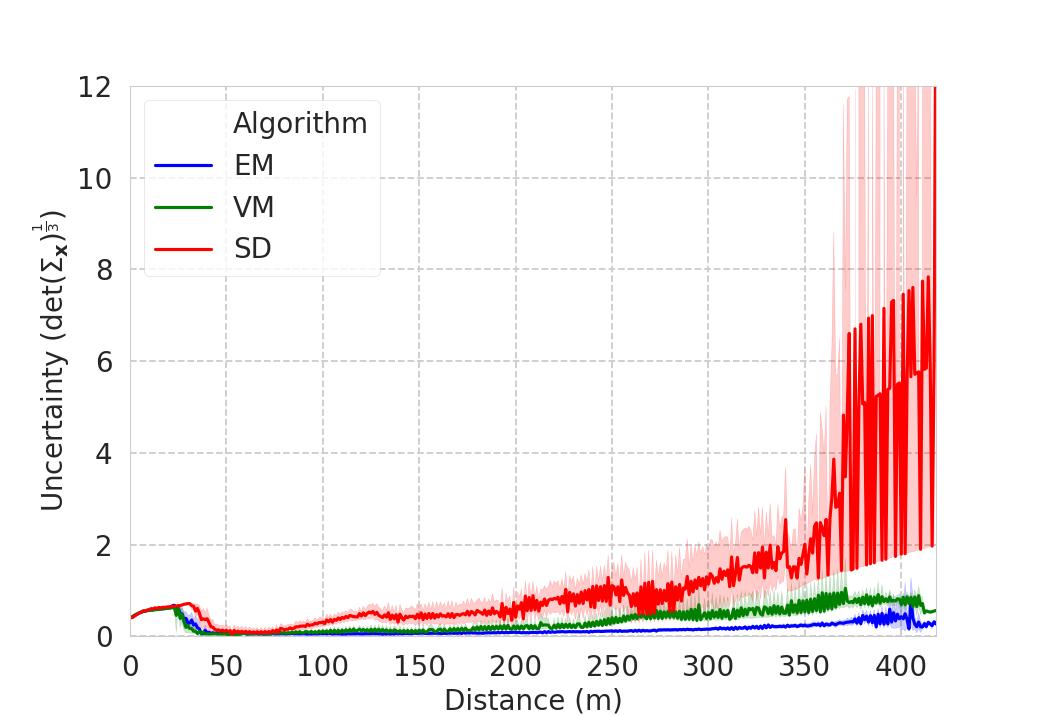}
    \label{fig:FF3}
\end{subfigure} 
\caption{\textbf{Fish farm results:} goals 1, 2, 3 shown from top to bottom.}
\label{fig:FF}
\vspace{-6mm}
\end{figure}

The virtual map for the fish farm environment with start (blue star) and goal locations (red cross) can be seen in Fig. \ref{fig:FF_VM}, and corresponding results can be observed in Fig. \ref{fig:FF}. Resembling the previous experiment, VM and EM achieve a similarly low uncertainty performance while SD uncertainty has a higher and less stable uncertainty result. Fig. \ref{fig:FF} (top) shows that for goal 1 the SD planner manages to reduce its pose error, before  pose error spikes up again. Likewise, Fig. \ref{fig:FF} (bottom) shows that for goal 3 the SD planner manages to lower the pose error after a spike, potentially due to a coincidental loop closure. However, EM and VM still achieve lower pose error than SD in all three scenarios, with EM being slightly superior. 

\subsubsection{Bridge-tunnel Environment}
\begin{figure}
\centering
\begin{subfigure}{0.5\textwidth}
    \includegraphics[width=0.5\textwidth]{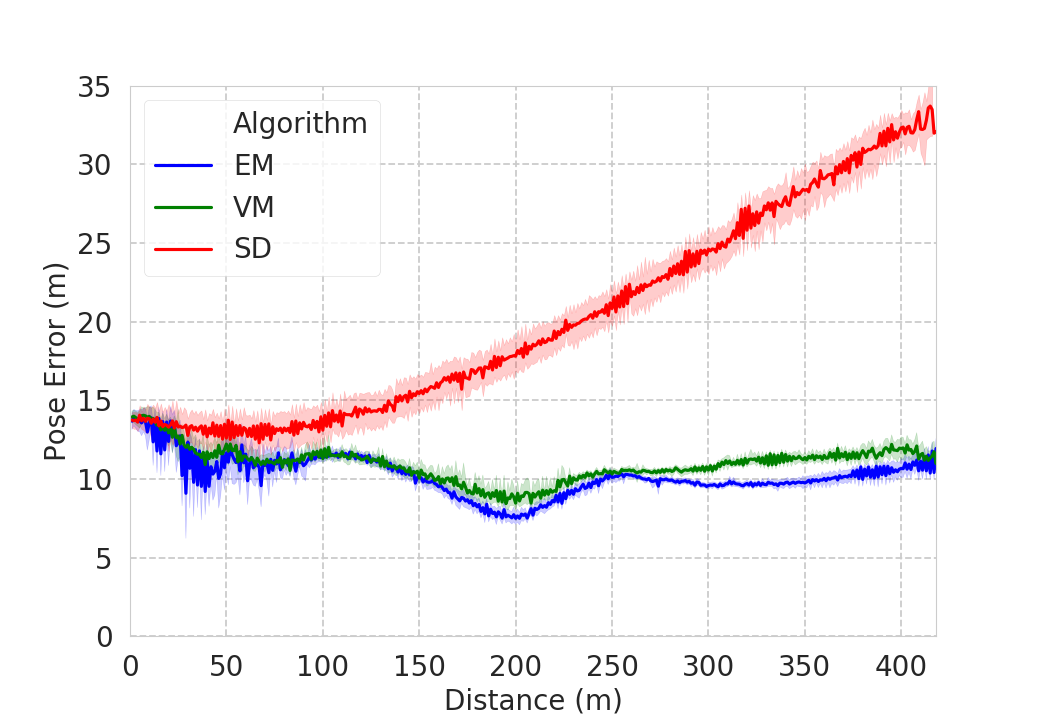}
    \hspace{-5mm}
    \includegraphics[width=0.5\textwidth]{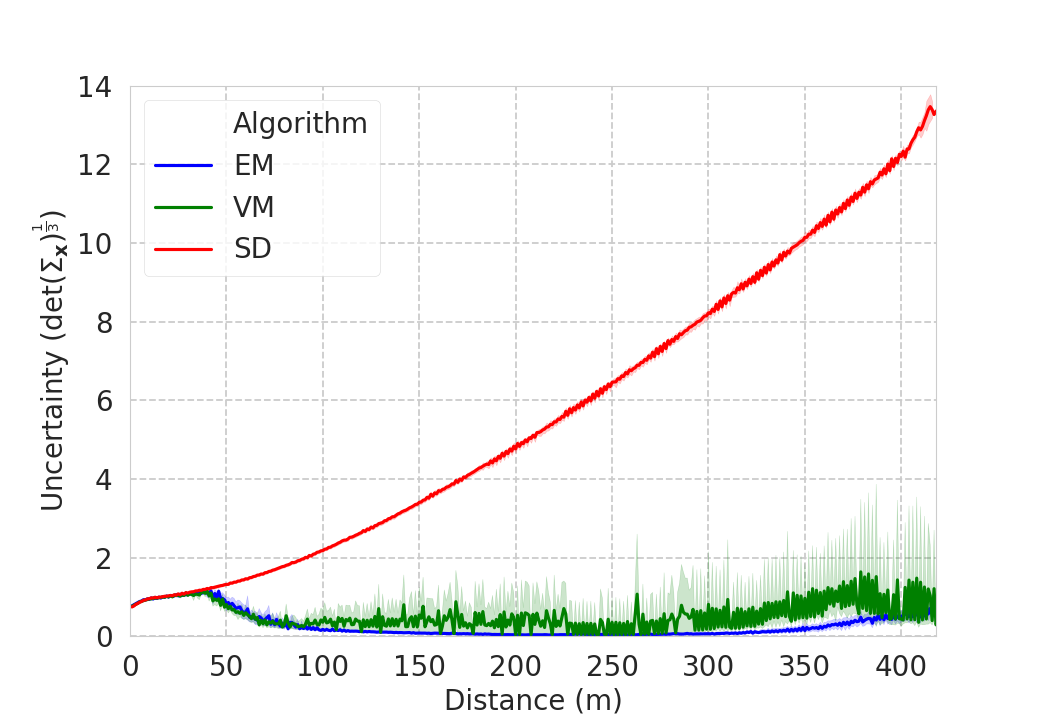}
    \label{fig:C1}
\end{subfigure}
\hspace{-5mm}
\begin{subfigure}{0.5\textwidth}
    \includegraphics[width=0.5\textwidth]{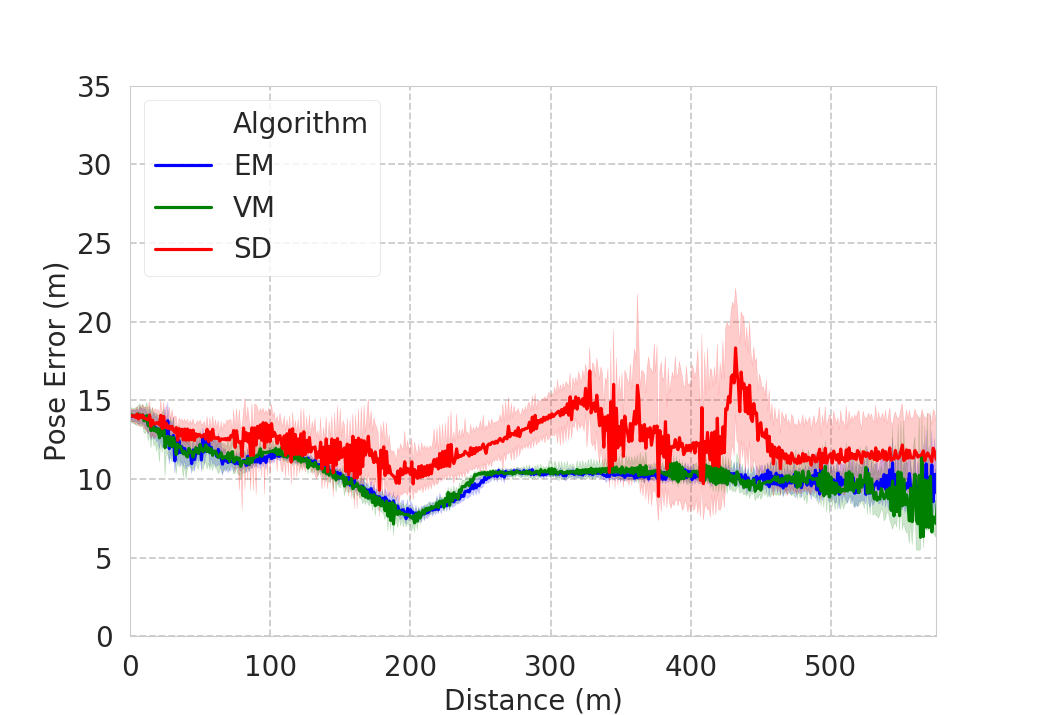}
    \hspace{-5mm}
    \includegraphics[width=0.5\textwidth]{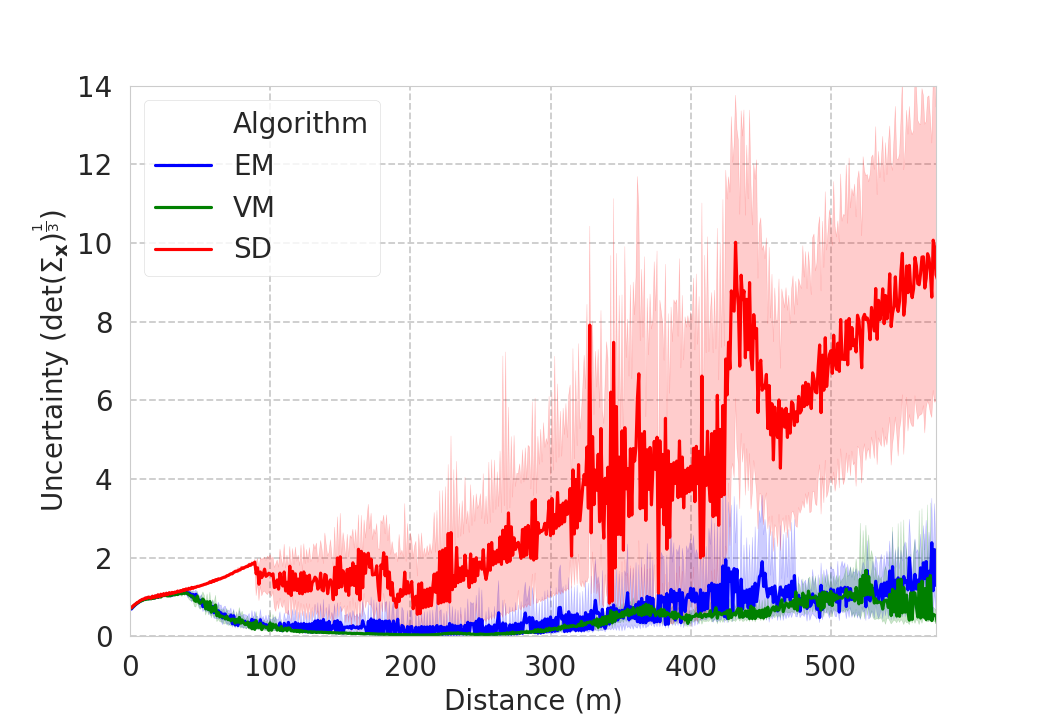}
    \label{fig:C2}
\end{subfigure}   
\hspace{-5mm}
\begin{subfigure}{0.5\textwidth}
    \includegraphics[width=0.5\textwidth]{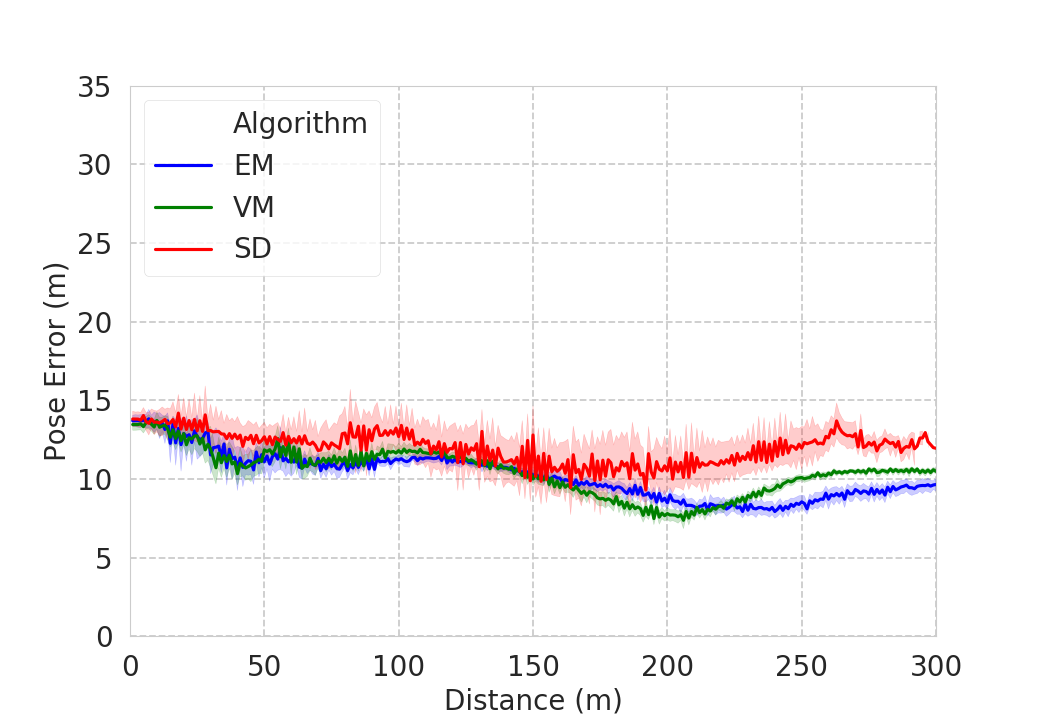}
    \hspace{-5mm}
    \includegraphics[width=0.5\textwidth]{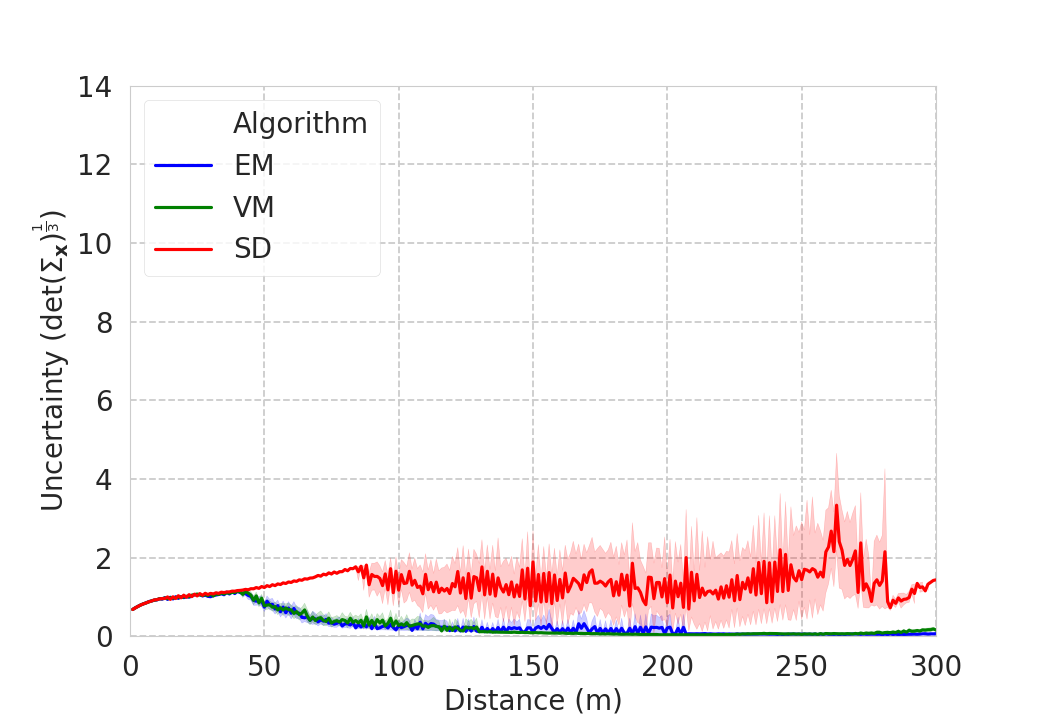}
    \label{fig:C3}
\end{subfigure} 
\caption{\textbf{Bridge-tunnel results:} goals 1, 2, 3 shown from top to bottom.}
\label{fig:C}
\vspace{-2mm}
\end{figure}

The virtual map and relevant locations for the bridge-tunnel environment are shown in Fig. \ref{fig:C_VM}, and results are shown in Fig. \ref{fig:C}. For all three targets, VM performance is similar to EM, while SD uncertainty grows unbounded and exhibits higher pose estimation errors. 

\subsubsection{Result Summary}
Fig. \ref{fig:example} shows example paths achieved by each planner, where the ellipses represent the evolution of uncertainty. It is clear that the VM and EM paths are similar in error and uncertainty; they both keep close to features in the environment and achieve low uncertainty by triggering SLAM loop closures. VM and EM perform drastically better lowering uncertainty than the SD planner. Lastly, Table \ref{tab:sec} shows the amount of time it takes to evaluate each candidate action and choose the appropriate desired path. In the SD planner, only one path is assessed and chosen. In our proposed VM planner and in the EM planner, 6 candidate actions are assessed each iteration. The results show the average time of every utility evaluation performed in all 100 trials. As expected, SD planner has the fastest computation time while EM has the slowest. It can be seen that EM takes 3 orders of magnitude more time than VM, while VM and SD maintain the same order of magnitude.

\begin{table}[tbp]
    \centering
    \footnotesize
        \setlength\tabcolsep{3pt}
      \begin{tabular}{|c|c c|c c|c c|}
      \hline
        \multicolumn{7}{|c|}{ \textbf{Computation Time (sec)}}\\
        \hline
        \multicolumn{1}{|c|}{\textbf{}}&\multicolumn{2}{|c|}{\textbf{EM}}&\multicolumn{2}{|c|}{\textbf{VM}}&\multicolumn{2}{|c|}{\textbf{SD}}\\
        
        \multicolumn{1}{|c|}{}&\multicolumn{1}{|c}{\text{($\mu$)}}&\multicolumn{1}{c|}{\text{($\sigma$)}}&\multicolumn{1}{|c}{\text{($\mu$)}}&\multicolumn{1}{c|}{\text{($\sigma$)}}&\multicolumn{1}{|c}{\text{($\mu$)}}&\multicolumn{1}{c|}{\text{($\sigma$)}}\\
        \hline
         {Marina 1}&{3.3553}&{0.1446}&{0.0041}&{0.0001}&{0.0015}& {0.0002}\\
         
        {Marina 2}&{2.9324}&{0.0598}&{0.0039}&{0.0002}&{0.0014}&{0.0002}\\
          
        {Marina 3}&{3.1507}&{0.1396}&{0.0040}&{0.0001}&{0.0015}&{0.0004}\\
          \hline
        {Fish Farm 1}&{5.4149}&{0.2094}&{0.0057}&{0.0001}&{0.0023}&{0.0001}\\
         
        {Fish Farm 2}&{4.9547}&{0.1768}&{0.0054}&{0.0001}&{0.0021}&{0.0001}\\
          
        {Fish Farm 3}&{5.4370}&{0.2138}&{0.0057}&{0.0001}&{0.0023}&{0.0001}\\
          \hline
        {Bridge-Tunnel 1}&{6.2699}&{0.0744}&{0.0069}&{0.0001}&{0.0027}&{0.0001}\\
         
        {Bridge-Tunnel 2}&{6.4998}&{0.1647}&{0.0070}&{0.0001}&{0.0027}&{0.0001}\\
          
        {Bridge-Tunnel 3}&{6.1624}&{0.1358}&{0.0067}&{0.0001}&{0.0026}&{0.0002}\\
          \hline
     \end{tabular}
     \caption{\textbf{Computation time comparison}: we give the mean and standard deviation of the utility evaluation time required in each experiment.}
     \label{tab:sec}
\vspace{-6mm}
\end{table}

\vspace{-1mm}
\section{Conclusion}
\label{sec:conclusion}
\vspace{-1mm}
This paper presents a computationally efficient planning under uncertainty framework for underwater robots operating in large-scale feature-sparse environments. We adapt the concept of virtual maps, a product of a robot's initial exploration of a new environment, using them as costmaps to avoid expensive online belief propagation while planning. A receding horizon motion planning strategy is implemented for planning to specified goals in a manner that is compatible with real AUV operations. The advantages of the proposed framework are demonstrated in a realistic underwater simulation. Results show a decrease in uncertainty and pose error compared to a standard shortest-distance approach. Furthermore, our approach is much faster than full belief propagation, while still maintaining low uncertainty and pose error. We aim for future work to extend this strategy to 3D planning scenarios, and to perform virtual map updates to accommodate important changes in the environment. 
\bibliographystyle{IEEEtran}
\bibliography{bib.bib}

\end{document}